\documentclass[sigconf]{acmart}

\renewcommand\footnotetextcopyrightpermission[1]{} 
\makeatletter
\renewcommand\@formatdoi[1]{\ignorespaces}
\makeatother
\acmConference[AI4FashionSC'20]{The fifth international workshop on fashion and KDD}{August 23, 2020}
{Virtual Event, CA, USA}

\usepackage{amsmath}
\usepackage{subfigure}
\usepackage{multicol}
\DeclareMathOperator*{\argmax}{arg\,max}

\AtBeginDocument{%
  \providecommand\BibTeX{{%
    \normalfont B\kern-0.5em{\scshape i\kern-0.25em b}\kern-0.8em\TeX}}}

\begin{document}
\title{Hyper-local sustainable assortment planning}

\author{Nupur Aggarwal}
\email{nupaggar@in.ibm.com}
\affiliation{%
  \institution{IBM Research}
}

\author{Abhishek Bansal}
\email{abansa27@in.ibm.com}
\affiliation{%
  \institution{IBM Research}
}

\author{Kushagra Manglik}
\email{kmangli1@in.ibm.com}
\affiliation{%
  \institution{IBM Research}
}

\author{Kedar Kulkarni}
\email{kdkulkar@in.ibm.com}
\affiliation{%
  \institution{IBM Research}
}

\author{Vikas Raykar}
\email{viraykar@in.ibm.com}
\affiliation{%
  \institution{IBM Research}
}

\renewcommand{\shortauthors}{Aggarwal, et al.}

\begin{abstract}
Assortment planning, an important seasonal activity for any retailer, involves choosing the right subset of products to stock in each store. While existing approaches only maximize the expected revenue, we propose including the environmental impact too, through the Higg Material Sustainability Index. The trade-off between revenue and environmental impact is balanced through a multi-objective optimization approach, that yields a Pareto-front of optimal assortments for merchandisers to choose from. Using the proposed approach on a few product categories of a leading fashion retailer shows that choosing assortments with lower environmental impact with a minimal impact on revenue is possible.
\end{abstract}




\maketitle

\section{Introduction}
While fashion is a multi-billion dollar global industry it comes with severe \textbf{environmental and social costs} worldwide. The fashion industry is considered to be the world’s second largest polluter, after oil and gas. Fashion accounts for 20 to 35 percent of microplastic flows into the ocean and outweighs the carbon footprint of international flights and shopping combined \cite{bof-article-2020-the-year-ahead}. Every stage in a garment's life threatens the planet and its resources. For example, it can take more than 20,000 liters of water to produce 1kg of cotton, equivalent to a single t-shirt and pair of jeans. Up to 8,000 different chemicals are used to turn raw materials into clothes, including a range of dyeing and finishing processes. This also has social costs with factory workers being underpaid and exposed to unsafe workplace conditions, particularly when handling materials like cotton and leather that require extensive processing \cite{mckinsey-article-2016-style}. Since fashion is heavily trend-driven and most retailers operate by season (for example, spring/summer, autumn/winter, holiday etc.), at the end of each season any unsold inventory is generally liquidated. While smaller retailers generally move the merchandise to second-hand shops, large brands resort to recycling or destroying the merchandise. 

In recent years this has led to addressing \textbf{sustainability} challenges as a core agenda for most fashion companies. Increasing pressure from investors, governments and consumer groups are leading to companies adopting sustainable practices to reduce their carbon footprint. Moreover, several companies may have sustainability targets (due to government regulations and/or self-imposed) to honor, which may lead to significant changes in the entire fashion supply/value chain. Sustainable practices can be adopted at various stages of the fashion value chain and several efforts are underway including more sustainable farming practices for growing fabric (for example, cotton), material innovation for alternatives (to cotton fabric, leather, dyes etc.), end-to-end transparency/visibility in the entire supply chain, sourcing from sustainable suppliers, better recycling technologies and sustainability index for measuring the full life-cycle impact of an apparel. In this paper, our main focus is to address sustainability challenges in the pre-season assortment planning activity in the fashion supply chain.

\textbf{Assortment planning} is a common pre-season planning done by buyers and merchandisers. Typically a fashion retailer has a large set of products under consideration to be potentially launched for the next up coming season. These could be a combination of \textit{existing products} from the earlier seasons along with the \textit{new products} that are designed for the next season. The designers interpret the fashion trends to design and develop a certain number of products for each category as specified in the option plan. The final products (both existing and new) are presented to the buyer and merchandiser who then curate/select a subset of them as the assortment for the next season. This assortment planning is typically based on her estimation of how well the product will sell (based on historical sales data and her interpretation of trends). During the initial planning the team works only with the initial designs or some times a sample procured from a vendor. Once the assortment has been selected the buyer then works with the sourcing team and the vendors to procure the products. The choice of the final assortment is a crucial decision since it has a big impact on the sell through rate, unsold inventory and eventually the revenue for the next season. 

In practice, the merchandiser has to actually select a different assortment for each region or store, referred to as, \textbf{hyper-local assortment planning}. While a retailer has a large set of products to offer, due to budget and space constraints only a smaller number of products can be stocked at each store. In this context, one of the most crucial planning tasks for most retail merchandisers is to decide the right assortment for a store, that is, what set of products to stock at each store.

The current practice for assortment planning is heavily spreadsheet driven and relies on the expertise and intuition of the merchandisers, coupled with trends identified from the past sales transactions data. While it is still manageable for a merchandiser to plan an assortment for a single store, it is not scalable when a merchandiser has to do planning for hundreds of stores. Typically stores are grouped into store clusters and an assortment is planned for each cluster rather than store. A sub-optimal assortment results in excess leftover inventory for the unpopular items, increasing the inventory costs, and stock outs of popular items, resulting in lost demand and unsatisfied customers. With better assortment planning algorithms retailer are now open to more algorithmic store-level automated assortments. 

The task of hyper-local assortment planning is to determine the optimal subset of products (from a larger set of products) to be stocked in each store so that the revenue/profit is maximized under various constraints and at the same time the assortment is localized to the preferences of the customers shopping in that store. The notion of a store can be generalized to a location and can potentially include store, region, country, channel, distribution center etc.

Existing approaches to assortment planning only maximize the expected revenue under certain store and budget constraints. Along with the revenue the choice of the final assortment has also an environmental cost associated with it. The final environmental impact of an assortment is eventually the sum of the environmental impact of each of the products in the assortment. In this paper, we address the notion of \textbf{sustainable assortments} and optimize the assortments under additional sustainability constraints. 

To achieve this we need a metric to measure the environmental impact of an apparel. One of the main deciding factors is the fabric or the kind of material used in the apparel. For example, cotton, accounting for about 30 percent of all textile fiber consumption, is usually grown using a lot of water, pesticides, and fertilizer, and making 1 kilogram of fabric generates an average of 23 kilograms of greenhouse gases. In this work we use the \textbf{Higg Material Sustainability Index} (MSI) score which is the apparel industry's most trusted tool to accurately measure the environmental sustainability impacts of materials \cite{higg-msi}. The Higg MSI score allows us to quantify the effect of using different materials, for example, while the cotton fabric has a score of 98, viscose/rayon fabric is a more sustainable fabric with score of 62. While we demonstrate our algorithms with the Higg MSI score any other suitable sustainability metric can be incorporated in our framework. 

While designers and merchandisers strive to make sustainable fabric choices during the design phase there is always a trade-off involved between sustainable choices and achieving high sell through rates. Also, the choice will typically be made at an individual product level and it is hard for the designer or buyer to assess the environmental impact of the assortment as a whole. The trade-off between revenue and environmental impact is balanced through a multi-objective optimization approach, that yields a Pareto-front of optimal assortments for merchandisers to choose from.

The rest of the paper is organized as follows. In \S~\ref{ref:assortment-planning} we define the problem of hyper-local assortment planning. In \S~\ref{ref:sustainability-scores}, we present the sustainability score calculations. In \S~\ref{ref:sustainable-assortment-planning}, we outline our approach to do a sustainable assortment planning. In \S~\ref{ref:experiments}, we present experimental results of our approach.

\section{Hyper-local assortment planning}
\label{ref:assortment-planning}
We define an (hyper-local) assortment for a store as a subset of $k$ products carried in the store from the total $n$ (potential) products. The task of \textbf{assortment planning} is to determine the optimal subset of $k$ products to be stocked in each store so that the \textit{assortment is localized to the preferences of the customers shopping in that store}. The optimization is done to maximize sales or gross margin subject to financial (limited budget for each store), store space (limited shelf life for displaying products) and other constraints.

Broadly there are three aspects to assortment planning, (1) the choice of the \textbf{demand model}, (2) \textbf{estimating the parameters} of the chosen demand model and (3) using the demand estimates in an \textbf{assortment optimization} setup.

\subsection{Demand Models}
The starting point for any assortment planning is to leverage an accurate demand forecast at a store level for a product the retailer is planning to introduce this season. The demand for a product is dependent on the assortment present in the store when the purchase was made. Several models have been proposed in the literature to model the demand. The forecast demand will then be used in a suitable stochastic optimization algorithm to do the assortment planning and refinement.

Given a set of $n$ substitutable products $\mathcal{N} = \{1,2,...,n\}$  and $m$ stores $\mathcal{S} = \{1,2,...,m\}$, let $d_{js}(\mathbf{q}_s)$ be the \textbf{demand} for product $j \in \mathcal{N}$ at store $s \in \mathcal{S}$ when the assortment offered at the store was $\mathbf{q}_s \subset \mathcal{N}$. An alternate construct is to specify it as a customer \textbf{choice} model $p_{js}(\mathbf{q}_s)$ which is the probability that a random customer chooses/prefers the product $j$ at store $s$ over other products in the assortment offered at the store. 

\textbf{Independent demand model} The simplest approach is to assume product demand to be independent of the offer set or the assortment, that is, the demand for a product does not depend on other available products. This model can therefore be specified by a discrete probability distribution over each of the products.

\begin{equation}
    p_{js}(\mathbf{q}_s) = \mu_{js} \quad \text{if } j \in \mathbf{q}_s \quad \text{such that} \sum_{j \in \mathcal{N}} \mu_{js} = 1
\end{equation}

This is the simplest demand model that has been traditionally around in retail operations, and assumes no substitution behavior. In practice the demand for a product is heavily influenced by the assortment that is under offer mainly due to \textbf{product substitution} (cannibalization) and \textbf{product complementarity} (halo-effect). The literature here is mainly focused on various parametric and non-parametric discrete choice models to capture product substitution, including, multinomial logit and variants\cite{kok-2008}, the exponomial discrete choice model\cite{alptekinoglu-2016}, deep neural choice models \cite{otsuka-2016} \cite{mottini-2017} and non-parametric rank-list models \cite{farias-2017}.

Since the main focus is to address the notion of sustainability in assortment for ease of exposition in this paper, we mainly focus on this simple independent demand model and ignore the effects of substitution. In general, any demand model can be plugged into the optimization framework.

\subsection{Estimating demand models}

Once an appropriate demand/choice model is chosen the parameters of the model have to be estimated based on historical sales and inventory data. Different demand models come with its own challenges and computation complexities in estimating the model parameters and include least squares, standard gradient based optimization, column generation and EM algorithms to maximize the likelihood. Berbeglia et al. 2019 \cite{berbeglia-2018} presents a good overview and a comparative empirical study of different choice-based demand models and their parameter estimation algorithms. 

For the independent demand model, we mainly rely on the historical store-level sales data to get an estimate of $d_{js}$ and multiply it by a suitable scalar to capture the trend increase or decrease for that year.

\begin{itemize}
    
\item  For existing products that were historically carried at a store, this is essentially the number of units of the products sold in the last season.

\item  However, in general, not all products are historically carried at all stores. For existing products that were not carried at the store, we use \textbf{matrix factorization} approaches to estimate the demand by modeling the problem as a product $\times$ store matrix and filling in the missing entries via matrix completion. This is described in more detail in Section \nameref{MF}.

\item  For completely new products without any previous sales history, then we use its visual and textual attributes to get a multi-modal embedding, and based on that we forecast the store-wise potential sales. \cite{ekambaram-kdd-2020}.

\end{itemize}

\subsection{Matrix Factorization} \label{MF}

Matrix factorization (MF) popularized in the collaborative filtering and recommender systems literature \cite{koren-2009} is commonly used to impute missing data. Let $\mathbf{X}$ be a $\texttt{product} \times \texttt{store}$ matrix of dimension $n \times m$ where each element $X_{ij}$ of the matrix represents the metric (for example, total sales) associated with product $i$ at store $j$. This matrix is sparse with elements missing for products not carried at the store. MF essentially decomposes this sparse matrix into two lower dimensional matrices $\mathbf{U}$ and $\mathbf{V}$ where  $\mathbf{U} \in \mathbb{R}^{n \times D}$ and $\mathbf{V} \in \mathbb{R}^{m \times D}$, such that rows in $\mathbf{U}$ and $\mathbf{V}$ encapsulate the product and store embeddings of dimension $D$. These $D$ dimensional embeddings (latent vectors) namely $\mathbf{U}_{i}$ and $\mathbf{V}_{j}$ are expected to capture the underlying hidden structure that influences the sales for product $i$ and store $j$ respectively. A common approach towards MF is to use Alternating Least Squares algorithm, however, other regularization extensions have also been characterized at length in the literature. In this paper, we have adopted the Alternating Least Squares approach and minimize the following loss function.
\begin{align}\begin{split}
\mathbf{L}(\mathbf{X},\mathbf{U},\mathbf{V}) = \sum_{i,j}c_{ij}(X_{ij} - \mathbf{U}_{i}\mathbf{V}_j^{T} - \beta_{i}-\gamma_{j})^{2} + \lambda(\sum_{i}(\|\mathbf{U}_{i}\|+\beta_{i})^{2} + \\ \sum_{j}(\|\mathbf{V}_{j}\|+\gamma_{j})^{2})
\end{split}\end{align}
where $\mathbf{\beta}$ and $\mathbf{\gamma}$ are product and store bias vectors of dimension $n$ and $m$ respectively and $c_{ij}$ be the weightage given to observed entries based on their upper and lower bounds limit. Once the loss function gets minimized we estimate the unseen entries $X_{ij}^{*}$, as follows.
\begin{equation}
    X_{ij}^{*} = \mathbf{U}_{i}\mathbf{V}_j^{T} + \beta_{i} + \gamma_{j}
\end{equation}
Thus, matrix $\mathbf{X}$ which was initially sparse now gets completely filled and is fed into our assortment planning module.

\subsection{Assortment optimization}

The forecast demand will then be used in a suitable stochastic optimization algorithm to do the assortment planning. The task of assortment optimization is to choose an optimal subset of products to maximize the expected revenue \textbf{subject to various constraints}.
\begin{equation}
\mathbf{q}_s^{*} = \argmax_{\mathbf{q}_s \subset \mathcal{N}} \sum_{j \in \mathbf{q}_s} \pi_{js} d_{js}(\mathbf{q}_s)
\end{equation}

where $\pi_{js}$ is the expected revenue when the product $j$ is sold at store $s$. Some of the commonly used constraints include,

\textbf{Cardinality constraints} The number of products to be included in an assortment is specified via a coarse range plan (sometimes also called an option plan or buy plan) for a store. The range plan specifies either the count of products or the total budget the retailer is planning to launch for a particular season as the granularity of category, brands, attributes and price points. 

\textbf{Diversity constraints} For some domain it is important to ensure that the selected assortment is \textit{diverse} to offer greater variety to the consumer. Without the diversity constraints the assortment tends to prefer products that are similar to each other. The general framework is to define a \textbf{product similarity function}  which measures the similarity between two products and use that as an additional constraint in the optimization.

\textbf{Complementarity constraints} The other important aspect is that a good assortment has products that are frequently bought together. Product complementarity (sometimes referred to as halo-effect) refers to behavior where a customer buys another product (say, a \textit{blue jeans}) that typically goes well with a chosen product (say, \textit{white top}). 

In this paper, we mainly focus on cardinality constraints. Our main contribution is to introduce environmental impact as additional constraints to the assortment optimization problem. As a result, this helps in making optimal assortment decisions in supply chains while accounting for both the economic and the environmental impact.


\section{Sustainability scores}
\label{ref:sustainability-scores}

We need a metric to measure the environmental impact of an apparel. One of the main deciding factors is the fabric or the kind of material used in the apparel.

We calculate the sustainability score for a product using the \textbf{Higg Material Sustainability Index (MSI)} developed by the Sustainable Apparel Coalition \cite{higg-msi}. The Higg MSI quantifies impact score for each fabric by taking into account various processes involved in the manufacturing of fabrics such as raw material procurement, yarn formation, textile formation, dyeing etc. Higg MSI calculates the impact on climate change, eutrophication, resource depletion, water scarcity and chemistry. The score is calculated for each impact area, then normalized followed by a weighted average. The Higg MSI score allows us to quantify the effect of using different materials; for example, while cotton has a score of 98, viscose/rayon is a more sustainable fabric with a score of 62. 

The Higg MSI value corresponds to consolidated environmental impact of 1 kg of a given material. Moreover, products made up of these constituent materials will typically have different weights. Thus, we adjust the Higg MSI of a product based on its weight.

For blended fabrics, we take a weighted average of Higg MSI of individual fabrics in the same proportions as they are in the blend. 

\begin{equation}
h_j = (\sum_{f \in F}H_f * p_f) \times w_j
\end{equation}

where $p_f$ is fabric percentage for each fabric $f$ present in the blended fabric $F$, $H_f$ is the Higg MSI for fabric $f$, $w_j$ is the weight of the product in $\texttt{kg}$, $h_j$ is the sustainability score for product $j$. For a set of $N$ products in an assortment, the sustainability score can be calculated as 
\begin{equation}
     h_{a_N} = \frac{1}{N} (\sum_{j \in \mathcal{N}}h_j )
\end{equation}

It should be noted that the Higg MSI is a cradle-to-gate index and doesn't consider downstream processes such as the impact due to laundry, wear and tear etc.

\section{Sustainable assortment planning}
\label{ref:sustainable-assortment-planning}

Once we have the store-wise product-wise demand/sales forecasts and the Higg MSI score for each product we can formulate this as a multi-objective optimization problem where we are interested in selecting those products for which we have better sales forecasts and at the same time that result in a sustainable assortment. Moreover, instead of just one solution, we would like to give the user a set of solutions near the Pareto Optimal front so that the user can visualize and select whichever assortment satisfies her criteria. We solve the following multi-objective problem for each store $s$.
\begin{equation}
\mathbf{x}_s^{*} = \argmax_{\mathbf{x}_s \in \{0,1\}^{n}, ||\mathbf{x}_s|| \le k} \frac{(1- \lambda)}{k} \underbrace{\sum_{j \in \mathcal{N}} \pi_{js} d_{js} x_{js}}_{\text{revenue}} - \frac{\lambda}{k}\underbrace{\sum_{j \in \mathcal{N}} h_j x_{js}}_{\text{sustainability}}
\end{equation}

where $x_{js}$ is a binary variable denoting presence or absence of product $j$ from the assortment at the store $s$, $d_{js}$ the demand for product $j$ at store $s$, $h_j$ is the weighted Higg MSI score for that product and $\lambda$ is a parameter through which the user can specify the relative importance of each objective. In the results section we show the optimal Pareto frontier by varying $\lambda$.

\subsection{Multi-objective optimization}

As described in the earlier section, the objective of the assortment planning problem is to determine optimal assortments that have the least Higg MSI score (least environmental impact) with a minimal impact on the sales. Optimizing these two objectives – maximizing sales and minimizing the Higg MSI score – individually will likely yield fundamentally different assortment solutions that may lead to superior sales but with a high Higg MSI (high environmental impact) or vice versa. To address this trade-off, formulating the assortment planning problem as a multi-objective optimization problem that optimizes the sales and the Higg MSI score at the same time is justified.

Multi-objective optimization problems have been formulated and solved using classical methods as well as meta-heuristics in literature \cite{deb-2001}. Of the available methods, the weighted sum method is employed to formulate and solve the assortment planning problem, due to its simplicity in configuration and use. In this method, relative importance of different objectives, as represented by multiplicative coefficients of the objective functions, is continually changed; and for each realization of these coefficients, a single-objective optimization problem is solved yielding an optimal assortment. Solving the single-optimization problem for multiple coefficient realizations yields a family of Pareto-optimal assortments that are non-dominated with respect to each other in the objective function space of sales and the Higg MSI score. The merchandiser can then choose from these optimal assortment solutions, depending on the preferred balance between sales and environmental impact.

In the proposed formulation, for a given $\lambda$, we compute the single objective function score for each product, which is a weighted combination of the sustainability and quality (revenue) scores. We then choose the top $k$ products as the assortment.





\section{Experimentation Validation}
\label{ref:experiments}

For our experimental validation, our main goals were to visualize the effect of including sustainability on the assortment. We used a dataset obtained from a leading fashion retailer consisting of 3484 products and sales over the time period Spring-Summer 2018 season. Product weight and the product's fabric composition were used to calculate the Higg MSI score for each product. We analyzed the \textit{upper} category that mainly consisted of \textit{t-shirts, shirts, tops} (total 1600 products). We calculated the Higg MSI score and the quality score (sales forecast) using the methods outlined in the previous sections.

\subsection{Sustainability and Quality Distribution}
\begin{figure}
\includegraphics[width=0.4\textwidth]{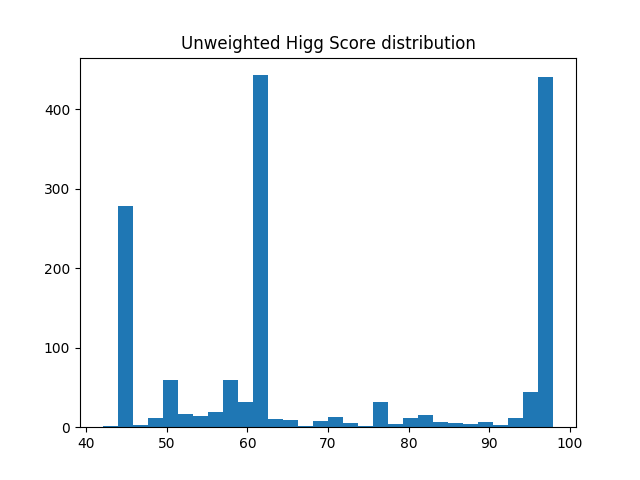}
\caption{Histogram distribution of product Higg MSI scores.}
\label{fig:higg-dist}
\end{figure}

\begin{figure}
\includegraphics[width=0.4\textwidth]{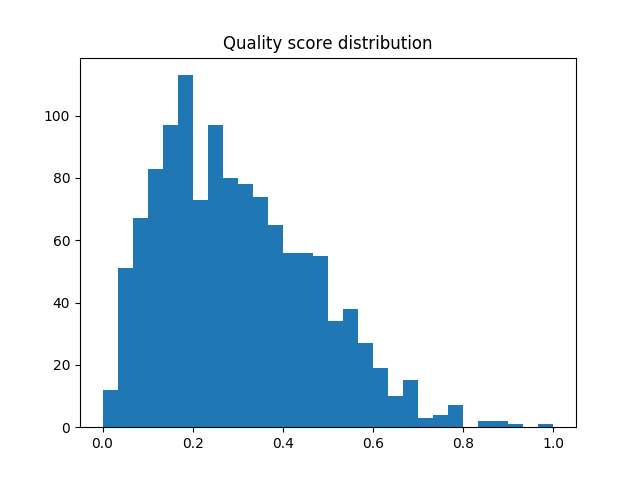}
\caption{Histogram distribution of product quality scores.}
\label{fig:quality-dist}
\end{figure}

Before planning the assortments, we visualized the distribution of different sustainability and quality scores that the products had (Figures \ref{fig:higg-dist}, \ref{fig:quality-dist}). In the plots we can see that the quality scores are evenly distributed; however, there are three peaks in the Higg MSI scores. On investigating further, we found that these corresponded to those products for which the fabric composition was either 100\% cotton or 100\% viscose or 100\% polyester where cotton had the highest Higg MSI (least sustainable) and polyester had the least Higg MSI (most sustainable).




\subsection{Pareto Front for Assortment Optimization}
\begin{figure}[]
  \centering
  \subfigure[Assortment size 1: All products are plotted.]{\includegraphics[scale=0.3]{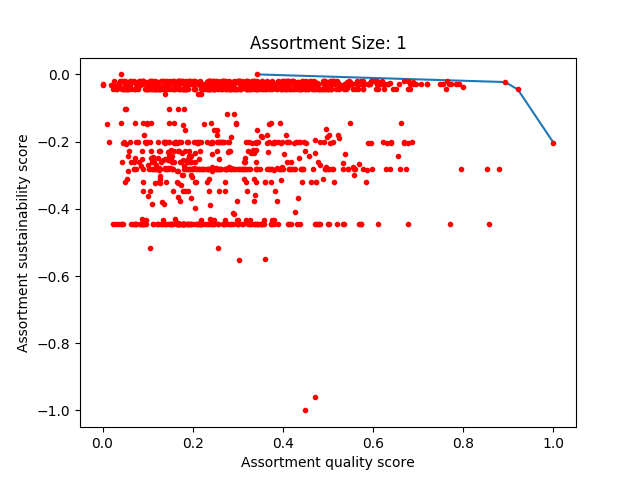}}
  \subfigure[Assortment size 10: All points on Pareto Front are plotted, besides 2000 randomly chosen assortments.]{\includegraphics[scale=0.3]{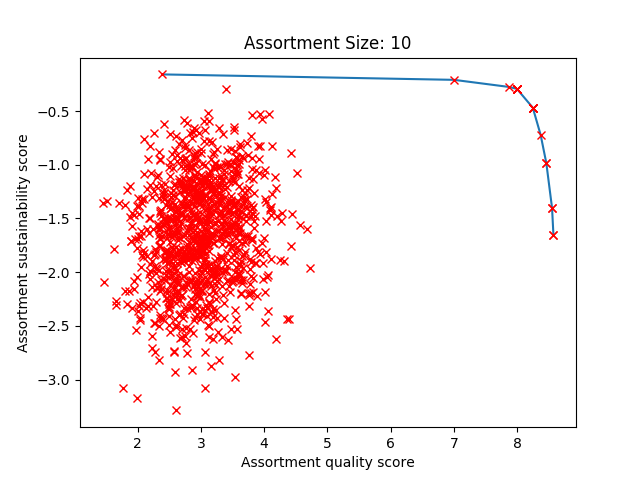}}
  \caption{Pareto Optimal fronts for varying assortment sizes.}
  \label{fig:pareto}
\end{figure}

We ran our optimization algorithm for multiple assortment sizes and plotted the Pareto optimal front by varying $\lambda$, the relative importance weight given to sustainability and revenue (quality) from 0 to 1. (Figure \ref{fig:pareto})

In the plots, the blue curve corresponds to the optimal Pareto frontier. We can see that as we increase the assortment size, the Pareto frontier and the assortment cluster shrinks \textit{relative to the frontier}. This is because as we aggregate the scores of more products, the consolidated scores move closer to their mean. Also, the 3 horizontal clusters in assortment size 1 plot is consistent with our observation that the Higg MSI score distribution also contains 3 peaks corresponding to 100\% cotton, 100\% viscose and 100\% polyester products respectively.


\subsection{Fabric composition variation}

\begin{figure}[!t]
  \centering
  \subfigure[Pareto optimal for $\lambda = 0.0$]{\includegraphics[scale=0.2]{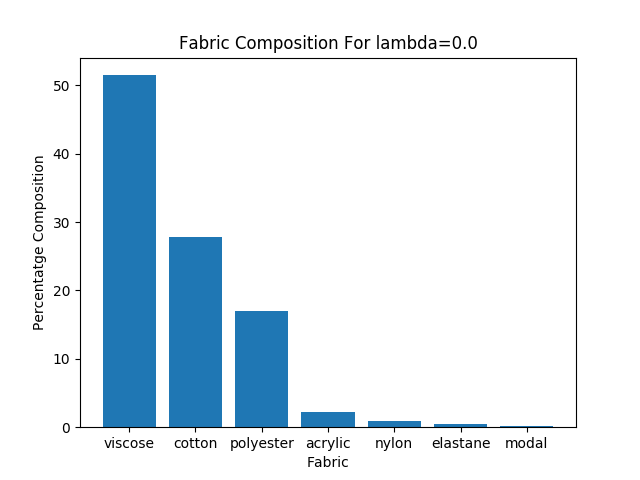}}
  \subfigure[Pareto optimal for $\lambda = 0.5$]{\includegraphics[scale=0.2]{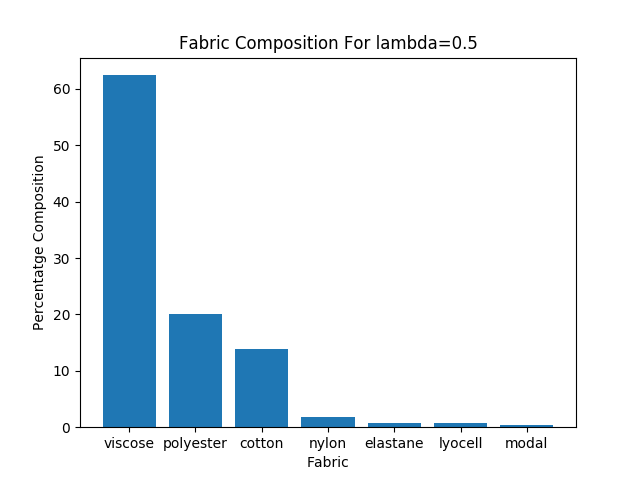}}
  \subfigure[Pareto optimal for $\lambda = 1.0$]{\includegraphics[scale=0.2]{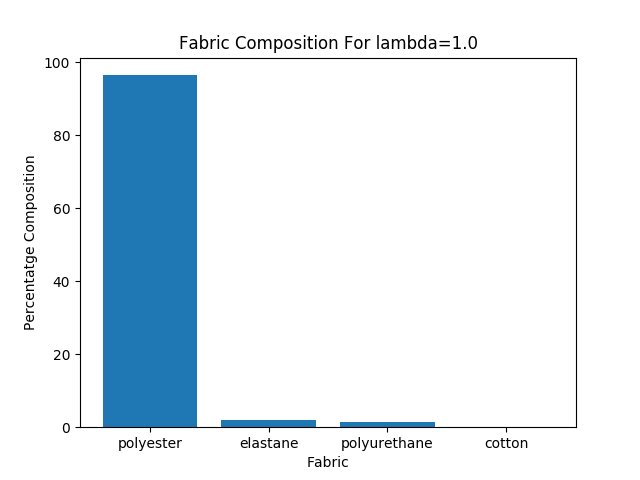}}
  \caption{Fabric Composition of extreme and middle points on the Pareto Optimal Frontier for assortment size 100.}
  \label{fig:fabric-comp}
\end{figure}

We further investigated and visualized the assortment compositions for 3 points on the Pareto Optimal frontier for assortment size 100, corresponding to $\lambda = 0.0, 0.5, 1.0$ and saw the interesting distributions plotted in Figure \ref{fig:fabric-comp}.


We can see that for $\lambda=1.0$ (maximum importance to sustainability), the fabric composition in the assortment products comprises mostly of polyester since its Higg MSI is the lowest signifying that it is most sustainable fabric. Looking at $\lambda=0.0$ and $\lambda=0.5$ plots we see that viscose fabric is dominant since its Higg MSI is lower than cotton and it has the best quality score in terms of quality scores as well.

\section{Conclusions and Future Work}
In this work, we have proposed a method of assortment planning that jointly optimizes the environmental impact of an assortment and the revenue. We formulated the problem as a multi- objective optimization problem whose optimal solutions lie on the Pareto Optimal front. The proposed approach would allow retailers to meet their sustainability targets with minimal impact on the revenue. In future work, we would like to consider cannibalization and halo effects in demand modeling as well. We would also like to consider diversity and complementarity of products in the assortment in the optimization formulation. Another extension would be to use a cradle-to-grave sustainability metric for assortment planning.


\bibliographystyle{ACM-Reference-Format}
\bibliography{sustainable-assortment}





\end{document}